\documentclass{article}
\usepackage{ijcai24}

\usepackage{times}
\usepackage{soul}
\usepackage{url}
\usepackage[hidelinks]{hyperref}
\usepackage[utf8]{inputenc}
\usepackage[small]{caption}
\usepackage{graphicx}
\usepackage{amsmath}
\usepackage{amsthm}
\usepackage{booktabs}
\usepackage{algorithm}
\usepackage{algorithmic}
\usepackage[switch]{lineno}
\usepackage{subcaption}
\usepackage{multirow} 

\title{GauU-Scene: A Scene Reconstruction Benchmark on Large Scale 3D Reconstruction Dataset Using Gaussian Splatting}

\author{
Butian Xiong $^1$
\and
Zhuo Li$^2$\and
Zhen Li$^3$
\affiliations
The Chinese University of Hong Kong, Shenzhen\\
\emails
\{butianxiong,zhuoli\}@link.cuhk.edu.cn,
zhenli@cuhk.edu.cn,
}

\begin{document}

\maketitle
\begin{figure*}[htbp]
  \centering
    \includegraphics[width=\textwidth]{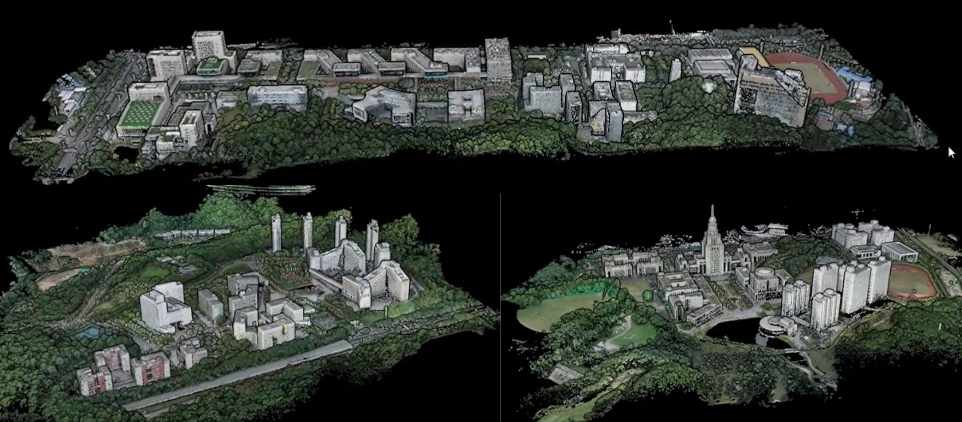}
   \caption{Our dataset is divided into three main parts. The first part is the top part of this graph. We call it CUHKSZ(The Chinese University of Hong Kong, Shenzhen) lower campus, and the bottom left corner shows the upper campus of CUHKSZ, and the bottom right corner shows the SMBU(Shenzhen MSU-BIT University) Campus. We use highly accurate lidar to collect the dataset and the range we cover is more than 1.5 $km^2$.}
   \label{fig:CUHKSZ_SMBU}
\end{figure*}

\begin{abstract}
We introduce a novel large-scale scene reconstruction benchmark using the newly developed 3D representation approach, Gaussian Splatting, on our expansive U-Scene dataset. U-Scene encompasses over one and a half square kilometres, featuring a comprehensive RGB dataset coupled with LiDAR ground truth. For data acquisition, we employed the Matrix 300 drone equipped with the high-accuracy Zenmuse L1 LiDAR, enabling precise rooftop data collection. This dataset, offers a unique blend of urban and academic environments for advanced spatial analysis convers more than 1.5 $km^2$. Our evaluation of U-Scene with Gaussian Splatting includes a detailed analysis across various novel viewpoints. We also juxtapose these results with those derived from our accurate point cloud dataset, highlighting significant differences that underscore the importance of combine multi-modal information.
\end{abstract}

\section{Introduction}
\label{sec:intro}
3D reconstruction is a transformative technology, enabling the conversion of real-world scenes into digital three-dimensional models. This technology, often involving the transformation of multiple 2D images into 3D models, fosters applications in urban planning, virtual reality (VR), and augmented reality (AR). Various techniques have been employed to enhance the accuracy and efficiency of 3D reconstruction. Among them, Structure from Motion (SfM) has been notably prominent, as extensively studied by \cite{snavely2006photo}. This photogrammetric technique uses 2D images captured from different angles to reconstruct three-dimensional structures.

A recent innovation in 3D reconstruction is the advent of Neural Radiance Fields (NeRF), which predicts volumetric scene representations from sparse 2D images using a fully connected neural network \cite{mildenhall2021nerf}. Despite initial challenges such as training difficulty and limitations in scale and complexity, rapid advancements have been made. Meta-learning techniques \cite{chen2021mvsnerf}, sparsity exploitation \cite{Zhang_2022_CVPR}, data structure integration \cite{martel2021acorn}, \cite{granskog2021neural}, and eigenvalue usage \cite{Chen2022ECCV} have significantly improved NeRF's performance. The current state-of-the-art method in addressing aliasing issues has also been proposed \cite{barron2022mip}.

NeRF models have been successful in synthesizing novel views of scenes, but the Gaussian Splatting Representation \cite{kerbl3Dgaussians} has emerged as a more recent 3D representation, combining rasterization with novel view synthesis for large-scale scenarios. While effective from a distance, Gaussian Splatting exhibits limitations such as blurring at closer inspection, which we analyze in detail in this study. We also explore the discrepancies between ground truth point cloud data and novel views generated by Gaussian Splatting. In the following section, we will focus on Gaussian Splatting and point out the drawback of Gaussian Splatting on the large-scale scene reconstruction.

Adapting different deep representation to large-scale environments, such as cityscapes, has been an emerging research area \cite{xiangli2022bungeenerf} \cite{blocknerf}. 

The challenge of city-scale reconstruction extends beyond the model to the dataset. Existing city-scale datasets \cite{xiangli2022bungeenerf} \cite{UrbanBIS} \cite{blocknerf} have limitations, either lacking accurate ground truth data or being focused on scene understanding rather than reconstruction. To address this, we employ the DJI Matrix 300 drone with Zenmuse L1 Lidar for capturing highly accurate 3D RGB point clouds. This approach enables us to venture beyond traditional indoor scenes, undertaking large city-scale outdoor scene reconstructions. Our dataset covers an area larger than 1.5 $km^2$, surpassing the scale of existing datasets like UrbanBIS \cite{UrbanBIS} and comparable to CityNeRF's RGB dataset \cite{xiangli2022bungeenerf}.

Additionally, we also provide a simple yet effective method to combine Lidar prior together with Gaussian-Splatting. The result by using this algorithm shows a boost to the accuracy of reconstruction result both quantitatively and qualitatively.

Our contributions are the following four:
\begin{itemize}
  \item Provision of a large dataset covering over 1.5 $km^2$ with a dense point cloud using Lidar as illustrated in the \ref{fig:CUHKSZ_SMBU}.
  \item Establishment of a concrete benchmark on our dataset using state-of-the-art Gaussian Splatting.
  \item Propose a Lidar-Image Fusion method that combine the prior obtained by lidar together with Gaussian Splatting to chonstruct a 3D Gaussian representation
  \item Shown the gap of drone based-large-scene reconstruction and future work 
\end{itemize}

In subsequent sections, we discuss related work in 3D Gaussian Splatting following by the preliminaries of how Gaussian Splatting works. After that we will introduce our measurement matrix, and detail our data collection methodology. Then we will discuss the method of transforming the lidar data to usable Gaussian Splatting Prior. Finally we will show the result and propose a potential gap in the current algorithm.

\begin{figure*}[htbp]
\centering
\includegraphics[width=\textwidth]{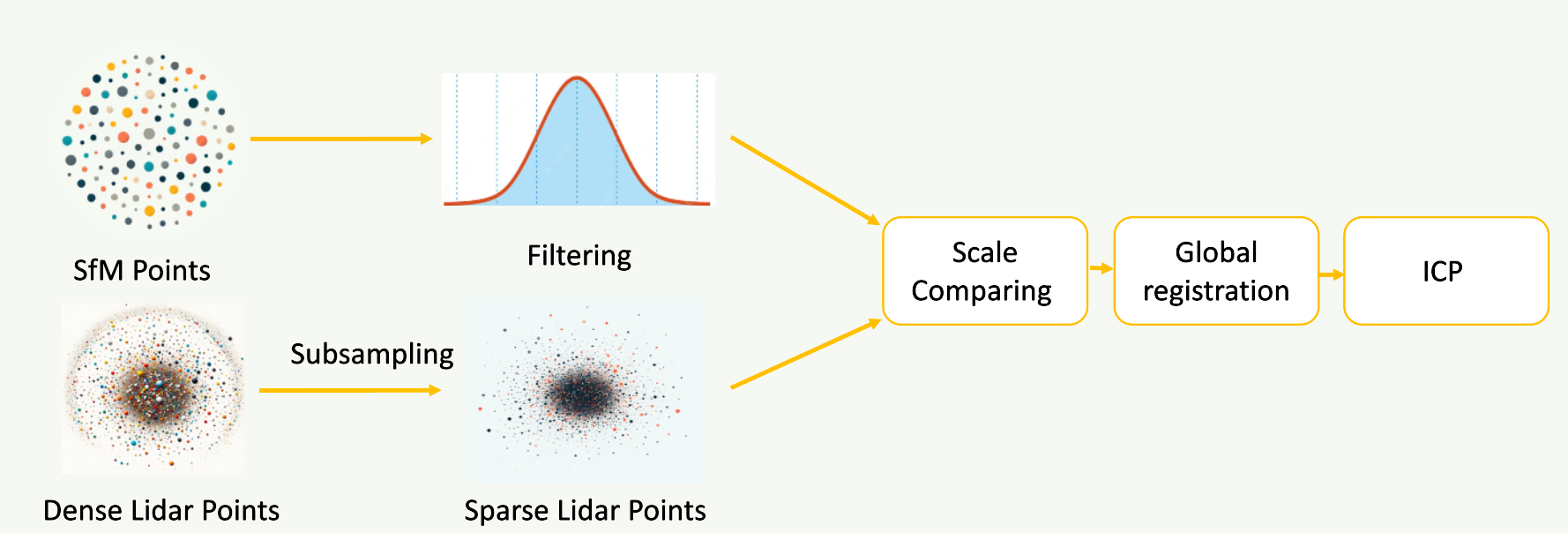}
\caption{The current point cloud registration method usually cannot handle different scales, so we first scale the raw point cloud to the same size as the SfM sparse point cloud. To do this, we find the maximum distance or variance in the SfM, as there are always some points far from the center in SfM. Then, we perform coarse matching manually and fine-tune it using ICP (Iterative Closest Point).}
\label{fig:preprocessing}
\end{figure*}
\section{Related Work}
In this section, we will first introduce the related dataset and their drawback and advantage for large scene reconstruction. Then we will introduce the emerged works of Gaussian Splatting including SLAM, Autonomous Driving, and Human-oriented Gaussian Splatting.

\subsection{Large Scale Dataset}
Outdoor large-scale datasets, such as \cite{xiangli2022bungeenerf}, are captured using different scale image data without ground truth. To evaluate their models, researchers typically select some novel views for generation and compare them with the collected data. However, using several images does not fully represent the entire 3D structure. Moreover, in \cite{xiangli2022bungeenerf}, they utilize varying scales from satellite-captured images, which exhibit clear differences in time and lack 3D ground truth. For instance, images captured in the morning and at night differ significantly, as do satellite images taken in 2010 and 2020 of the same region. We refer to this issue as the 'image time difference' problem in our current work. Another study, \cite{blocknerf}, attempts to provide ground truth point cloud data using Lidar mounted on cars and addresses the image time difference problem through style transformation. However, a major limitation of this approach is the lack of Lidar and photographic data from rooftops. \cite{UrbanBIS} employs drones with Lidar to capture point cloud data. As drones are fast and can collect data from rooftops, this method inherently solves the image-time-difference and rooftop issues mentioned earlier. Nevertheless, a critical drawback of this approach is the unclear correlation between Lidar point clouds and images due to differences in the coordinates of the camera and point cloud. Consequently, we can only use either point cloud or image data from \cite{UrbanBIS}. In our current work, we address these challenges and provide a clear, easy-to-follow pipeline to generate larger datasets that can utilize both point cloud and image data.

\subsubsection{Gaussian Splatting}

"Since the emergence of Gaussian Splatting \cite{kerbl3Dgaussians}, numerous studies have utilized 3D Gaussian Splatting (3DGS) as the primary representation for 3D models, encompassing both human figures and environments. While some research has sought to identify inherent issues with Gaussian Splatting, such as aliasing \cite{aliasing_3dgs}, many others have focused on downstream tasks using Gaussian Splatting. These include applications in autonomous driving environments \cite{drivegaussian}\cite{yan2024street}\cite{gsSLAM} and human representation \cite{dreamGauZiwei}\cite{gauhumanZiwei}\cite{humanGauZiwei}\cite{GPSyebin}\cite{animatableyebin}\cite{headAvataryebin}. However, there has been limited focus on employing Gaussian Splatting with drone-collected data. This current work is the first to attempt integrating Gaussian Splatting with a drone-based dataset."

\section{Preliminaries}
In this section, we introduce the testing method Gaussian Splatting we used in the current work. 

\subsection*{Gaussian Distribution}
A Gaussian distribution, often referred to as a normal distribution, is a continuous probability distribution characterized by a bell-shaped curve. It is completely determined by its mean $\mu$ and variance $\sigma^2$ (or standard deviation $\sigma$). The probability density function (PDF) of a Gaussian distribution is given by:
\begin{equation}
    p(x | \mu, \sigma^2) = \frac{1}{\sqrt{2\pi\sigma^2}}\exp\left(-\frac{(x - \mu)^2}{2\sigma^2}\right).
\end{equation}

In the multivariate case, the Gaussian distribution can be extended to accommodate a vector of means and a covariance matrix. The multivariate Gaussian PDF with mean vector $\boldsymbol{\theta}$ and covariance matrix $\boldsymbol{\Sigma}$ is:
\begin{equation}
    p(\mathbf{y} | \boldsymbol{\theta}, \boldsymbol{\Sigma}) = \frac{1}{\sqrt{(2\pi)^k |\boldsymbol{\Sigma}|}} \exp\left(-\frac{1}{2}(\mathbf{y} - \boldsymbol{\theta})^T \boldsymbol{\Sigma}^{-1}(\mathbf{y} - \boldsymbol{\theta})\right),
\end{equation}
where $|\boldsymbol{\Sigma}|$ denotes the determinant of $\boldsymbol{\Sigma}$, and $k$ is the dimensionality of the vector $\mathbf{y}$.

\subsection*{Covariance Matrix}
The covariance matrix $\boldsymbol{\Sigma}$ is a symmetric positive semi-definite matrix that contains the covariances between pairs of elements of the random vector. The covariance between two random variables $X$ and $Y$ is calculated as:
\begin{equation}
    \sigma_{XY} = \frac{1}{n-1} \sum_{i=1}^{n} (X_i - \overline{X})(Y_i - \overline{Y}),
\end{equation}
where $\overline{X}$ and $\overline{Y}$ are the sample means of $X$ and $Y$, respectively.

\subsection*{Symmetric Property and Optimization of the Covariance Matrix}
The eigenvalues and eigenvectors of the covariance matrix play a crucial role in understanding the shape and orientation of the distribution. The eigenvector corresponding to the largest eigenvalue points in the direction of the greatest variance (the major axis of the ellipse in the case of a bivariate Gaussian), with the eigenvalue indicating the magnitude of this variance. To maintain the positive semi-definite property of the covariance matrix during optimization, it is essential to parameterize the matrix in such a way that this property is not violated.

\subsection*{Proofs and Properties}
Additional proofs and properties of the Gaussian distribution and its covariance matrix are beyond the scope of this preliminary section, we provide the proof it in the appendix \ref{sec:Appendix}
\begin{figure*}[htbp!]
  \centering
  \includegraphics[width=\textwidth]{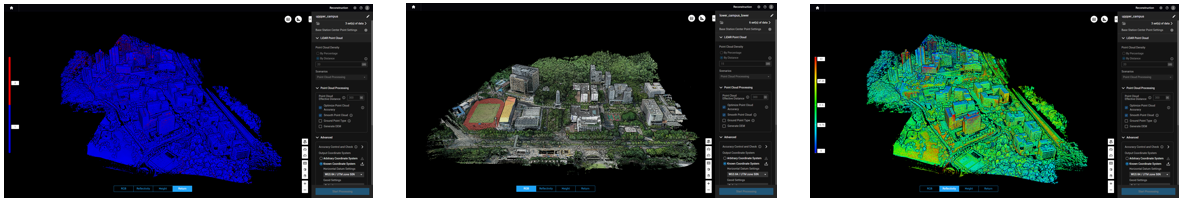} 
  \caption{The left one is the quality of the point if the point is blue, then the quality is ok, otherwise it is red. The right-hand side picture shows the point's altitude, and the middle one is the RGB point cloud}
  \label{fig:point_cloud}
\end{figure*}
\subsection{3D Representation}
In the context of 3D graphics, Gaussian Splatting is a technique used for rendering purposes, which involves representing points in space with Gaussian functions. Each point is represented as a Gaussian splat with several attributes that define its appearance in the rendered image. These attributes include:

\begin{itemize}
    \item $\mathbf{m}$: the mean vector, which defines the center of the splat in the 3D space.
    \item $\boldsymbol{\sigma}$: the covariance matrix, which determines the spread of the splat in different directions.
    \item $\alpha$: the opacity, which defines the transparency level of the splat.
    \item $\mathbf{h}$: the color, represented using Harmonic Spheres to encode the color variation within the splat.
\end{itemize}

A Gaussian splat can thus be defined by the tuple $<\mathbf{m}, \boldsymbol{\sigma}, \alpha, \mathbf{h}>$. The equation for a Gaussian splat incorporating these attributes is given by:

\begin{equation}
    G(\mathbf{m}, \boldsymbol{\sigma}, \alpha, \mathbf{h})
\end{equation}
when we do the rendering, we will integrate through the Gaussian Splatting in the 3D space using alpha blendering

The color function $\mathbf{h}(\mathbf{x})$ is defined on the Harmonic Spheres, which allow for a smooth transition of colors across the surface of the splat, adding a rich visual texture to the rendered scene.

\section{Method}

In this section, we will first introduce our raw data collection method. To construct a usable dataset, it is necessary to solve the problem of aligning coordinates, which will be introduced in the second subsection. Finally, the multi-modal training method will be discussed in the last subsection.

\subsection{Raw Data Collection}
The data collection technique mainly contains the following three stages. 
\begin{itemize}
  \item Drone assembly and hardware preparation. The detailed information is hard to describe in purely text, therefore, we provide a link to help our readers to understand the \href{https://lacy-backbone-098.notion.site/Drone-Assembly-864cdca6917e46cea145e0026f9096c5?pvs=4}{procedure}.
  \item Path Planning: Since we need to obtain all views of each building instead of just the rooftop, we need to set the routing algorithm in DJI-Pilot to the oblique shooting and select Zenmuse L1 and the camera, the detailed information can be examined in the oblique shooting \href{https://lacy-backbone-098.notion.site/Oblique-Shooting-0f59e1096c444c0687082b67b28aeb6a?pvs=4}{guidance}. 
  \item Flying: We need to use our controller to initiate fly according to the routing path we set before. There are something one needs to be very careful about, otherwise, there might be a crushing drone. For example, we need to keep keep distance, and altitude, we need to measure the wind speed and so on, the detailed information can be examined in the \href{https://lacy-backbone-098.notion.site/DJI-Drone-Operation-Tips-616a6f91481245e69997e89c6a5ba012?pvs=4}{tips}.
\end{itemize}

The equipment we use is the DJI Matrix 300 Flying platform together with Zenmuse L1 LiDAR. The detailed parameters are shown in the DJI official website: \href{https://enterprise.dji.com/matrice-300/specs}{Matrix300} and \href{https://enterprise.dji.com/zenmuse-l1/specs}{ZenmuseL1}.

\subsection{Data Pre-Processing}
The raw point cloud coordinates we collected are in the UTM system, provided by DJI TERRA. However, determining the camera position of the Zenmuse-L1 is challenging due to the rotation of both the gimbal and the drone, and the discrepancies between these two systems and the UTM coordinates. Therefore, instead of matching the camera position with the raw point cloud, we first create a sparse point cloud using COLMAP \cite{colmap1}\cite{colmap2}, simultaneously calibrating the camera position. We then register the sparse point cloud with the raw point cloud through scaling, global matching, and ICP (Iterative Closest Point) matching, converting the raw point cloud into relative coordinates in COLMAP. As shown in Fig.\ref{fig:preprocessing}, we first filter the SfM and downsample it. We then use global registration in CloudCompare or manually coarsely register the point cloud. Finally, we apply ICP to obtain the transformation matrix of the raw point cloud.

\subsection{Lidar-Image Fusion}
After converting the point cloud to the COLMAP coordinates, we can treat this point cloud as the initial point cloud for Gaussian Splatting. However, it is important to note that Gaussian Splatting requires a sparse point cloud as input, while the overall raw point cloud is too dense. Therefore, we need to subsample the point cloud to make it suitable for the network. By doing this, we successfully merge the Lidar point cloud with image information.

\section{Results}
In this section, we introduce the dataset size and qualitative results. The qualitative results can be examined on the following link to our project \href{https://saliteta.github.io/CUHKSZ_SMBU}{page}. The dataset is now published and available at this OneDrive \href{https://cuhko365-my.sharepoint.com/personal/120090584_link_cuhk_edu_cn/_layouts/15/onedrive.aspx?id=%2Fpersonal%2F120090584%5Flink%5Fcuhk%5Fedu%5Fcn%2FDocuments%2FDeepBit&ga=1}{link}, or it can be accessed through our project page. Moreover, we provide several videos showcasing the final results of Gaussian Splatting.

We demonstrate the results on our dataset using both Vanilla Gaussian Splatting and Lidar-Fused Gaussian Splatting. The key metrics we use are PSNR (Peak Signal to Noise Ratio) and L1 norm for image difference, and we employ the L1 norm to measure the differences in the point cloud and Gaussian Splatting. We present benchmarks exclusively on our dataset, as this current work is the first to combine drone-based point cloud data with Gaussian Splatting.

\subsection{Dataset}\
The dataset we provided here contains more important information such as the altitude and quality analysis result shown here in the figure \ref{fig:point_cloud}

\begin{figure}[ht]
    \centering
    \begin{subfigure}{.43\textwidth}
        \centering
        \includegraphics[width=\linewidth]{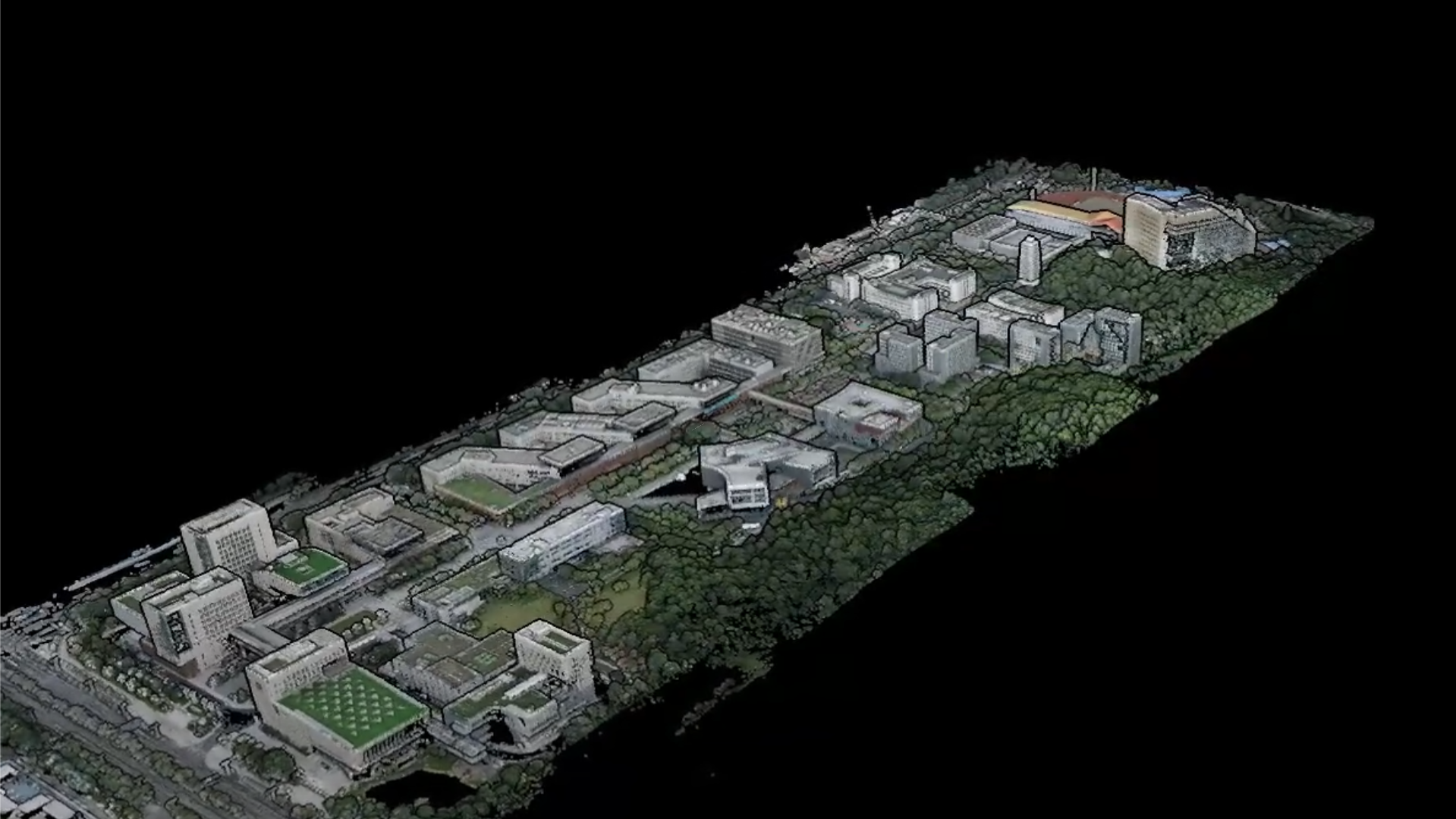}
        \caption{lower campus}
        \label{fig:sub1}
    \end{subfigure}
    \hfill
    \begin{subfigure}{.43\textwidth}
        \centering
        \includegraphics[width=\linewidth]{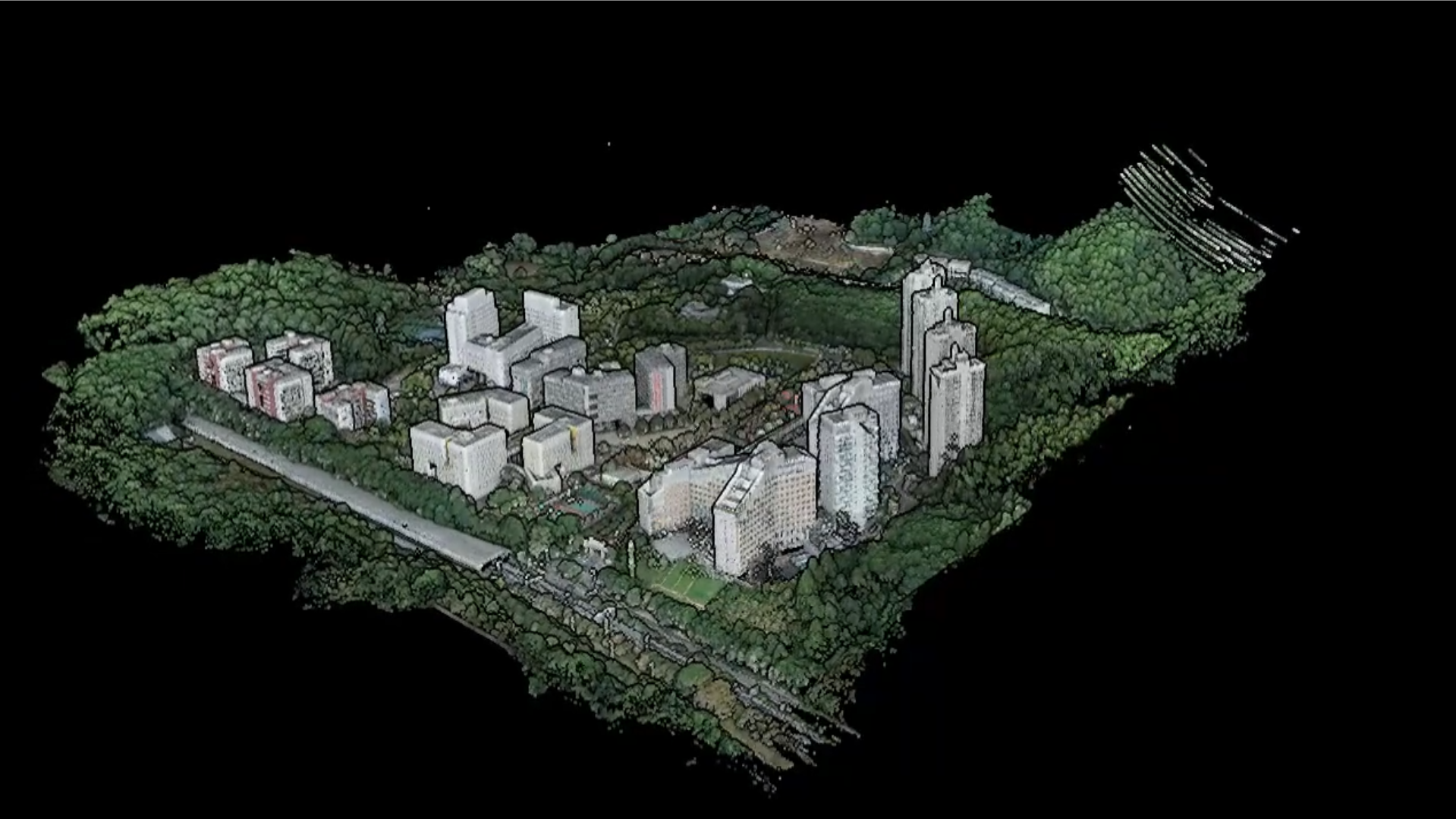}
        \caption{upper campus}
        \label{fig:sub2}
    \end{subfigure}
    \hfill
    \begin{subfigure}{.43\textwidth}
        \centering
        \includegraphics[width=\linewidth]{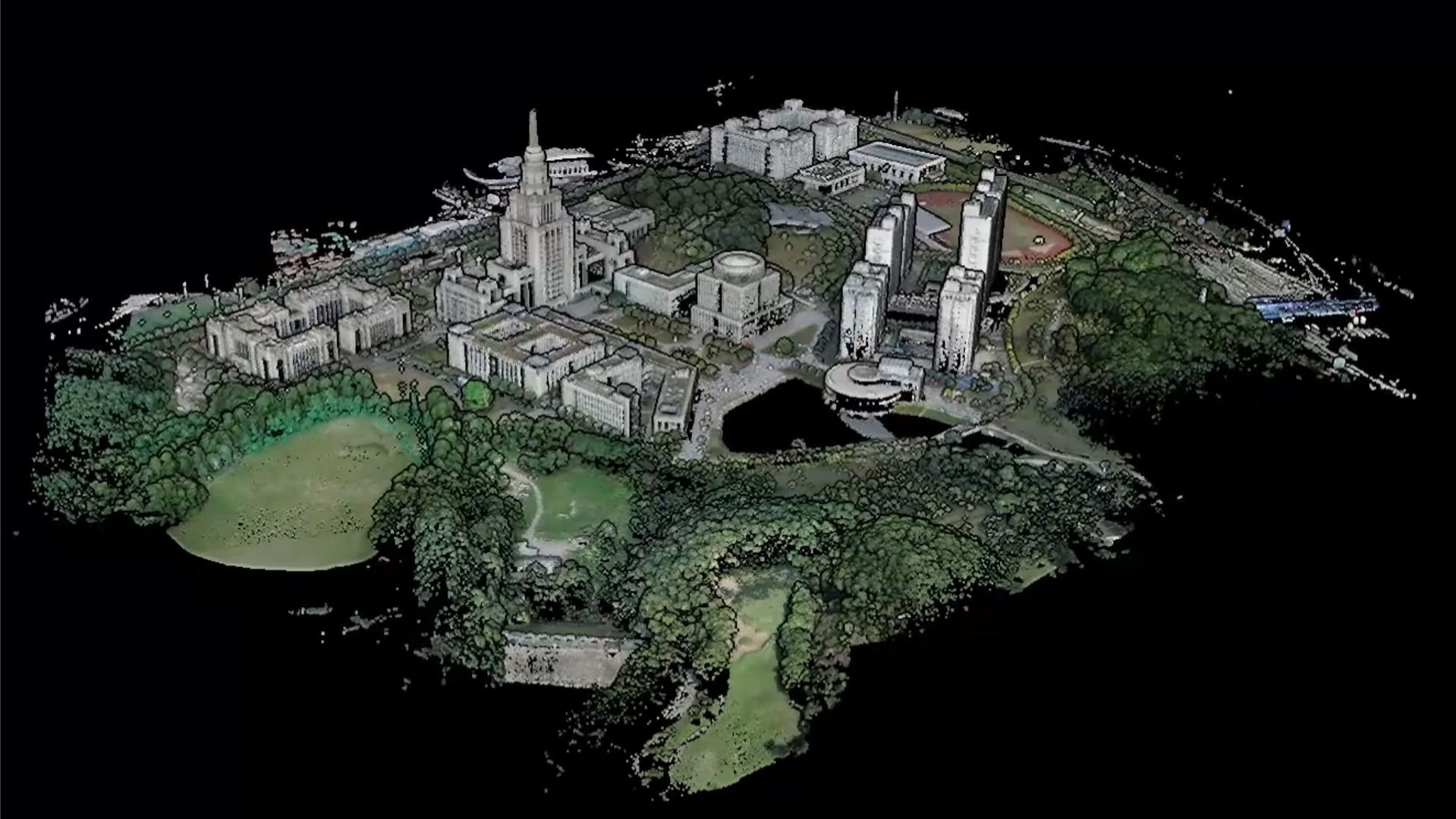}
        \caption{SMBU campus}
        \label{fig:sub3}
    \end{subfigure}
    \caption{The three figures here give another angle for our raw point cloud dataset}
    \label{fig:dataset}
\end{figure}

The point cloud data is in PLY format, and we sampled the final output to be 20cm per dot. The sample output looks like what shown in the \ref{fig:point_cloud}. The drone is flying at the altitude 120 meters, the effective point is withing 300 meters. The overall dataset covers more than 1.5 $km^2$. More visualized result shown in the Fig.\ref{fig:dataset}

\subsection{Lidar-Fused Gaussian Splatting}
\begin{figure*}[htbp]
  \centering
    \includegraphics[width=\textwidth]{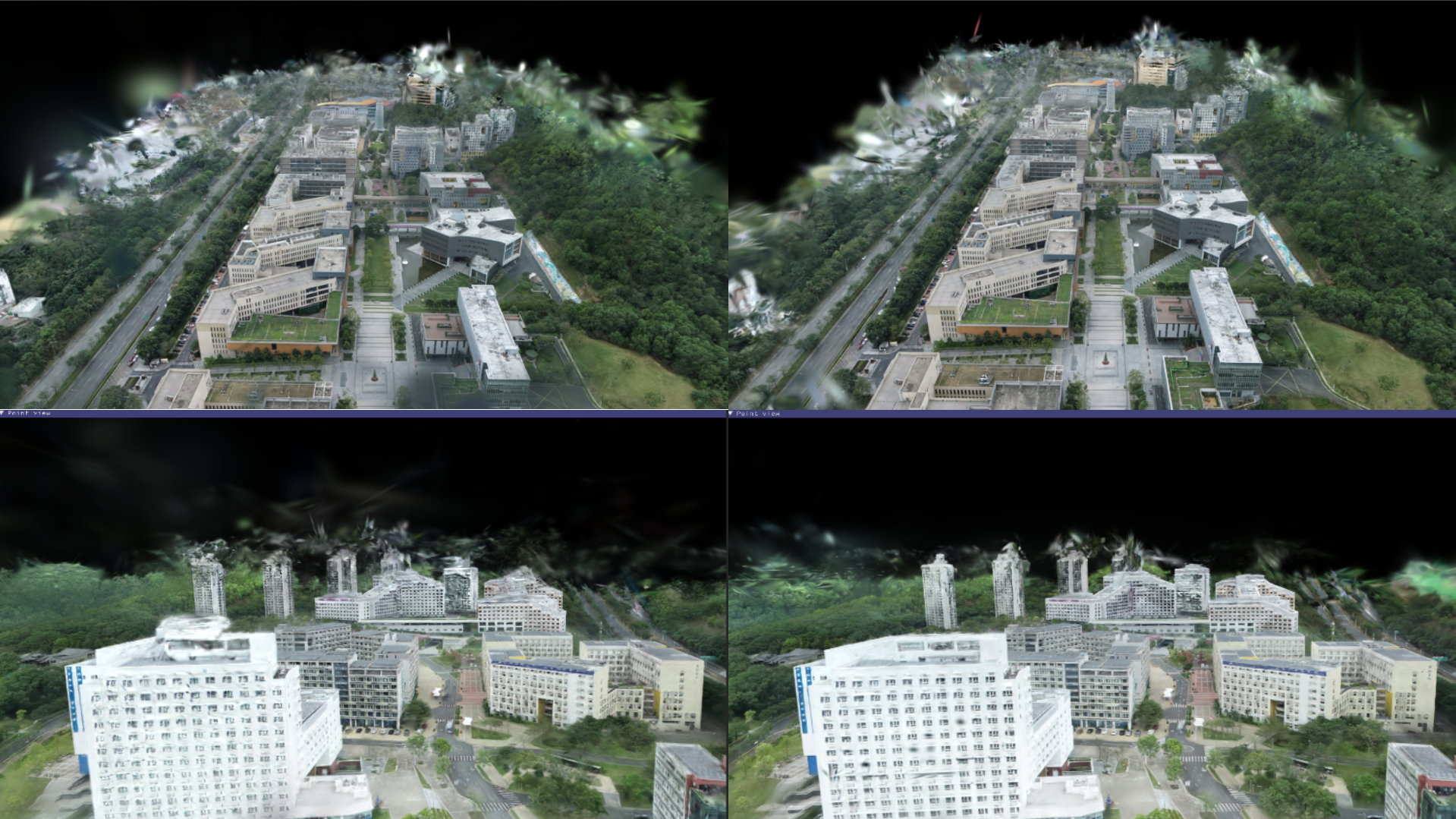}
   \caption{The left hand-side of this picture shows the result of using vanilla Gaussian Splatting, while the right hand side shows the result of using lidar-fused Gaussian Splatting. One can easily find the left hand-side pictures contains a some blurry black cloud, and also has an irregular blur building, while the right handside does not contain this defect}
   \label{fig:vanilavsGaussian}
\end{figure*}

In this section, we will introduce our fused algorithm result on our dataset. We first utilize L1 and PSNR as the measurement matrix. And we test our result only using the image.  
The result shown in the table.\ref{tab:image_result}

\begin{table}
    \centering
    \begin{tabular}{lrrr} 
        \toprule
        Dataset Name   & Method           & L1 Loss  & PSNR \\ 
        \midrule
        CUHK LOWER     & Lidar-Fused      & 0.0246   & 28.742   \\ 
                       & Vanilla          & 0.0250   & 28.660 \\ 
        CUHK UPPER     & Lidar-Fused      & 0.0321   & 26.949 \\ 
                       & Vanilla          & 0.0327   & 26.911 \\ 
        SMBU           & Lidar-Fused      & 0.0318   & 27.333 \\ 
                       & Vanilla          & 0.0321   & 27.010 \\ 
        \bottomrule
    \end{tabular}
    \caption{As one can find that the result shown cannot show much superiority of Lidar-Fusing algorithm. This is due to image cannot represent a precise 3D structure. }
    \label{tab:image_result}
\end{table}

As can be observed from the table, there is no significant difference between using Lidar-Fused prior and sparse SfM as a prior. While utilizing Lidar-Fused prior generally results in greater accuracy than using SfM for initialization, the degree of improvement is relatively minor. This outcome may be attributed to two factors: firstly, an image may not effectively represent a 3D object, and secondly, the metrics are calculated using the training images. Currently, every image is essential for constructing our 3D models.

In the second phase, we shifted our ground truth from images to point clouds. Each center of Gaussian Splatting is considered as a point, and we employ the Hungarian Maximum Matching Algorithm to find the closest point in the ground truth point cloud and determine the color difference between them. This difference is calculated using an L1 metric, and the subsequent table indicates a significant enhancement when using Lidar-Prior. We believe this is because a 3D point cloud can more accurately reflect the details of 3D models, whereas images may not coherently represent the 3D structure.

\begin{table}
    \centering
    \begin{tabular}{lrrr} 
        \toprule
        Dataset Name   & Method           & L1 Loss   \\ 
        \midrule
        CUHK LOWER     & Lidar-Fused      & 23.65     \\ 
                       & Vanilla          & 27.40    \\ 
        CUHK UPPER     & Lidar-Fused      & 23.43    \\ 
                       & Vanilla          & 27.69    \\ 
        SMBU           & Lidar-Fused      & 22.22    \\ 
                       & Vanilla          & 25.33    \\ 
        \bottomrule
    \end{tabular}
    \caption{As one can find that the result shown cannot show much superiority of Lidar-Fusing algorithm. This is due to image cannot represent a precise 3D structure. }
    \label{tab:point cloud result}
\end{table}

The qualitative result can be checked in the Fig.\ref{fig:vanilavsGaussian}

\section{Conclusion and Future Work}
In the current study we propose a dataset that solve the image time-difference problem and collect the rooftop information inherently utilizing the drone. We also provide a simple yet effective approach to solve the coordinates difference problem in lidar and images. Finally, we successfully combine the nature image and lidar information to feed as a prior to Gaussian Splatting. Our results shows a clear boost by fusing lidar and camera both qualitatively and quantitatively. And the difference between the matrix measure using the image ground truth and point cloud ground truth validate the neccesity of collecting point cloud ground truth. 
However, the dataset is still relatively small compare to \cite{UrbanBIS}, and there is no extra images for testing the result. Furthermore, the edge effect of Gaussian Splatting is clear. That is the edge of the 3D models is the main error that generated. Removing them in a smart way will effectively decrease the error and make a more reliable dataset.

\section{Appendix}
\label{sec:Appendix}
\subsection*{Proof of Eigenvalue Properties for Covariance Matrix}
In the context of the covariance matrix $\boldsymbol{\Sigma}$, we consider the eigendecomposition where $\boldsymbol{\Sigma} = \mathbf{Q}\boldsymbol{\Lambda}\mathbf{Q}^T$. Here, $\mathbf{Q}$ is the matrix of eigenvectors and $\boldsymbol{\Lambda}$ is the diagonal matrix of eigenvalues. 

Consider a pair of eigenvalues and eigenvectors $(\mathbf{x}_i, \lambda_i)$ of $\boldsymbol{\Sigma}$ with the property that $\lambda_i \geq \lambda_j$ for $i < j$. We aim to show that the largest eigenvalue corresponds to the direction of the greatest variance in the data, and subsequent eigenvalues correspond to orthogonal directions of decreasing variance.

\subsubsection*{Proof for the Longest Axis}
Given a unit vector $\mathbf{y}$, we want to show that it lies within the ellipse defined by the covariance matrix if $\mathbf{y}^T\boldsymbol{\Sigma}\mathbf{y} \leq \lambda_1$, where $\lambda_1$ is the largest eigenvalue. To show this, we consider that $\mathbf{y}$ can be expressed as a linear combination of the eigenvectors of $\boldsymbol{\Sigma}$:

\begin{equation}
    \mathbf{y} = n_1\mathbf{x}_1 + n_2\mathbf{x}_2 + \ldots + n_k\mathbf{x}_k \quad \text{with} \quad \sum n_i^2 = 1.
\end{equation}

Then we have:

\begin{equation}
    \mathbf{y}^T\boldsymbol{\Sigma}\mathbf{y} = \sum n_i^2\lambda_i \leq \lambda_1.
\end{equation}

\subsubsection*{Proof for Perpendicular Axes}
For the second largest eigenvalue $\lambda_2$, we consider the case where the vector corresponding to $\lambda_2$ is orthogonal to that of $\lambda_1$. We use the fact that $\mathbf{x}_1^T\boldsymbol{\Sigma}\mathbf{x}_2 = 0$ due to the orthogonality of the eigenvectors. Extending this to the entire set of eigenvectors, we can show that:

\begin{equation}
    \mathbf{x}_i^T\boldsymbol{\Sigma}\mathbf{x}_j = 0 \quad \text{for} \quad i \neq j.
\end{equation}

Hence, we conclude that the eigenvalues of the covariance matrix define the length of the axes of the ellipse (in the case of bivariate Gaussian) or ellipsoid (in the multivariate case), and the eigenvectors define their orientation.
\newpage
\bibliographystyle{named}
\bibliography{ijcai24}

\begin{thebibliography}{}

\bibitem[\protect\citeauthoryear{Barron \bgroup \em et al.\egroup }{2022}]{barron2022mip}
Jonathan~T Barron, Ben Mildenhall, Dor Verbin, Pratul~P Srinivasan, and Peter Hedman.
\newblock Mip-nerf 360: Unbounded anti-aliased neural radiance fields.
\newblock In {\em Proceedings of the IEEE/CVF Conference on Computer Vision and Pattern Recognition}, pages 5470--5479, 2022.

\bibitem[\protect\citeauthoryear{Chen \bgroup \em et al.\egroup }{2021}]{chen2021mvsnerf}
Anpei Chen, Zexiang Xu, Fuqiang Zhao, Xiaoshuai Zhang, Fanbo Xiang, Jingyi Yu, and Hao Su.
\newblock Mvsnerf: Fast generalizable radiance field reconstruction from multi-view stereo.
\newblock In {\em Proceedings of the IEEE/CVF International Conference on Computer Vision}, pages 14124--14133, 2021.

\bibitem[\protect\citeauthoryear{Chen \bgroup \em et al.\egroup }{2022}]{Chen2022ECCV}
Anpei Chen, Zexiang Xu, Andreas Geiger, Jingyi Yu, and Hao Su.
\newblock Tensorf: Tensorial radiance fields.
\newblock In {\em European Conference on Computer Vision (ECCV)}, 2022.

\bibitem[\protect\citeauthoryear{Granskog \bgroup \em et al.\egroup }{2021}]{granskog2021neural}
Jonathan Granskog, Till~N Schnabel, Fabrice Rousselle, and Jan Nov{\'a}k.
\newblock Neural scene graph rendering.
\newblock {\em ACM Transactions on Graphics (TOG)}, 40(4):1--11, 2021.

\bibitem[\protect\citeauthoryear{Hu and Liu}{2023}]{gauhumanZiwei}
Shoukang Hu and Ziwei Liu.
\newblock Gauhuman: Articulated gaussian splatting from monocular human videos.
\newblock {\em arXiv preprint arXiv:2312.02973}, 2023.

\bibitem[\protect\citeauthoryear{Kerbl \bgroup \em et al.\egroup }{2023}]{kerbl3Dgaussians}
Bernhard Kerbl, Georgios Kopanas, Thomas Leimk{\"u}hler, and George Drettakis.
\newblock 3d gaussian splatting for real-time radiance field rendering.
\newblock {\em ACM Transactions on Graphics}, 42(4), July 2023.

\bibitem[\protect\citeauthoryear{Li \bgroup \em et al.\egroup }{2023}]{animatableyebin}
Zhe Li, Zerong Zheng, Lizhen Wang, and Yebin Liu.
\newblock Animatable gaussians: Learning pose-dependent gaussian maps for high-fidelity human avatar modeling.
\newblock {\em arXiv preprint arXiv:2311.16096}, 2023.

\bibitem[\protect\citeauthoryear{Liu \bgroup \em et al.\egroup }{2023}]{humanGauZiwei}
Xian Liu, Xiaohang Zhan, Jiaxiang Tang, Ying Shan, Gang Zeng, Dahua Lin, Xihui Liu, and Ziwei Liu.
\newblock Humangaussian: Text-driven 3d human generation with gaussian splatting.
\newblock {\em arXiv preprint arXiv:2311.17061}, 2023.

\bibitem[\protect\citeauthoryear{Martel \bgroup \em et al.\egroup }{2021}]{martel2021acorn}
Julien N.~P. Martel, David~B. Lindell, Connor~Z. Lin, Eric~R. Chan, Marco Monteiro, and Gordon Wetzstein.
\newblock Acorn: {Adaptive} coordinate networks for neural scene representation.
\newblock {\em ACM Trans. Graph. (SIGGRAPH)}, 40(4), 2021.

\bibitem[\protect\citeauthoryear{Matsuki \bgroup \em et al.\egroup }{2023}]{gsSLAM}
Hidenobu Matsuki, Riku Murai, Paul~HJ Kelly, and Andrew~J Davison.
\newblock Gaussian splatting slam.
\newblock {\em arXiv preprint arXiv:2312.06741}, 2023.

\bibitem[\protect\citeauthoryear{Mildenhall \bgroup \em et al.\egroup }{2021}]{mildenhall2021nerf}
Ben Mildenhall, Pratul~P Srinivasan, Matthew Tancik, Jonathan~T Barron, Ravi Ramamoorthi, and Ren Ng.
\newblock Nerf: Representing scenes as neural radiance fields for view synthesis.
\newblock {\em Communications of the ACM}, 65(1):99--106, 2021.

\bibitem[\protect\citeauthoryear{Sch\"{o}nberger and Frahm}{2016}]{colmap1}
Johannes~Lutz Sch\"{o}nberger and Jan-Michael Frahm.
\newblock Structure-from-motion revisited.
\newblock In {\em Conference on Computer Vision and Pattern Recognition (CVPR)}, 2016.

\bibitem[\protect\citeauthoryear{Sch\"{o}nberger \bgroup \em et al.\egroup }{2016}]{colmap2}
Johannes~Lutz Sch\"{o}nberger, Enliang Zheng, Marc Pollefeys, and Jan-Michael Frahm.
\newblock Pixelwise view selection for unstructured multi-view stereo.
\newblock In {\em European Conference on Computer Vision (ECCV)}, 2016.

\bibitem[\protect\citeauthoryear{Snavely \bgroup \em et al.\egroup }{2006}]{snavely2006photo}
Noah Snavely, Steven~M Seitz, and Richard Szeliski.
\newblock Photo tourism: exploring photo collections in 3d.
\newblock In {\em ACM siggraph 2006 papers}, pages 835--846. 2006.

\bibitem[\protect\citeauthoryear{Tancik \bgroup \em et al.\egroup }{2022}]{blocknerf}
Matthew Tancik, Vincent Casser, Xinchen Yan, Sabeek Pradhan, Ben Mildenhall, Pratul~P Srinivasan, Jonathan~T Barron, and Henrik Kretzschmar.
\newblock Block-nerf: Scalable large scene neural view synthesis.
\newblock In {\em Proceedings of the IEEE/CVF Conference on Computer Vision and Pattern Recognition}, pages 8248--8258, 2022.

\bibitem[\protect\citeauthoryear{Tang \bgroup \em et al.\egroup }{2023}]{dreamGauZiwei}
Jiaxiang Tang, Jiawei Ren, Hang Zhou, Ziwei Liu, and Gang Zeng.
\newblock Dreamgaussian: Generative gaussian splatting for efficient 3d content creation.
\newblock {\em arXiv preprint arXiv:2309.16653}, 2023.

\bibitem[\protect\citeauthoryear{Xiangli \bgroup \em et al.\egroup }{2022}]{xiangli2022bungeenerf}
Yuanbo Xiangli, Linning Xu, Xingang Pan, Nanxuan Zhao, Anyi Rao, Christian Theobalt, Bo~Dai, and Dahua Lin.
\newblock Bungeenerf: Progressive neural radiance field for extreme multi-scale scene rendering.
\newblock In {\em European conference on computer vision}, pages 106--122. Springer, 2022.

\bibitem[\protect\citeauthoryear{Xu \bgroup \em et al.\egroup }{2023}]{headAvataryebin}
Yuelang Xu, Benwang Chen, Zhe Li, Hongwen Zhang, Lizhen Wang, Zerong Zheng, and Yebin Liu.
\newblock Gaussian head avatar: Ultra high-fidelity head avatar via dynamic gaussians.
\newblock {\em arXiv preprint arXiv:2312.03029}, 2023.

\bibitem[\protect\citeauthoryear{Yan \bgroup \em et al.\egroup }{2024}]{yan2024street}
Yunzhi Yan, Haotong Lin, Chenxu Zhou, Weijie Wang, Haiyang Sun, Kun Zhan, Xianpeng Lang, Xiaowei Zhou, and Sida Peng.
\newblock Street gaussians for modeling dynamic urban scenes.
\newblock {\em arXiv preprint arXiv:2401.01339}, 2024.

\bibitem[\protect\citeauthoryear{Yang \bgroup \em et al.\egroup }{2023}]{UrbanBIS}
Guoqing Yang, Fuyou Xue, Qi~Zhang, Ke~Xie, Chi-Wing Fu, and Hui Huang.
\newblock Urbanbis: a large-scale benchmark for fine-grained urban building instance segmentation.
\newblock In {\em SIGGRAPH}, pages 1--11, 2023.

\bibitem[\protect\citeauthoryear{Yu \bgroup \em et al.\egroup }{2023}]{aliasing_3dgs}
Zehao Yu, Anpei Chen, Binbin Huang, Torsten Sattler, and Andreas Geiger.
\newblock Mip-splatting: Alias-free 3d gaussian splatting.
\newblock {\em arXiv preprint arXiv:2311.16493}, 2023.

\bibitem[\protect\citeauthoryear{Zhang \bgroup \em et al.\egroup }{2022}]{Zhang_2022_CVPR}
Xiaoshuai Zhang, Sai Bi, Kalyan Sunkavalli, Hao Su, and Zexiang Xu.
\newblock Nerfusion: Fusing radiance fields for large-scale scene reconstruction.
\newblock In {\em Proceedings of the IEEE/CVF Conference on Computer Vision and Pattern Recognition (CVPR)}, pages 5449--5458, June 2022.

\bibitem[\protect\citeauthoryear{Zheng \bgroup \em et al.\egroup }{2023}]{GPSyebin}
Shunyuan Zheng, Boyao Zhou, Ruizhi Shao, Boning Liu, Shengping Zhang, Liqiang Nie, and Yebin Liu.
\newblock Gps-gaussian: Generalizable pixel-wise 3d gaussian splatting for real-time human novel view synthesis.
\newblock {\em arXiv preprint arXiv:2312.02155}, 2023.

\bibitem[\protect\citeauthoryear{Zhou \bgroup \em et al.\egroup }{2023}]{drivegaussian}
Xiaoyu Zhou, Zhiwei Lin, Xiaojun Shan, Yongtao Wang, Deqing Sun, and Ming-Hsuan Yang.
\newblock Drivinggaussian: Composite gaussian splatting for surrounding dynamic autonomous driving scenes.
\newblock {\em arXiv preprint arXiv:2312.07920}, 2023.

\end{thebibliography}

\end{document}